\documentclass[10pt,twocolumn,letterpaper]{article}

\usepackage{cvpr}              

\usepackage{graphicx}
\usepackage{amsmath}
\usepackage{amssymb}
\usepackage{booktabs}

\usepackage{csquotes}

\usepackage{multirow}
\usepackage{graphicx}
\usepackage{booktabs}
\usepackage{bbm}

\usepackage[accsupp]{axessibility}

\usepackage{color, colortbl}

\newcommand*{\twoelementtable}[3][l]%
{%
\renewcommand{\arraystretch}{0.8}%
\begin{tabular}[t]{@{}#1@{}}%
#2\tabularnewline
#3%
\end{tabular}%
}
\usepackage{array}
\usepackage{tabularx}

\usepackage{algorithm}
\usepackage{algorithmic}
\usepackage{multirow}
\usepackage{amsmath,amssymb}
\usepackage{xcolor} 

\newtheorem{Definition}{Definition}
\newtheorem{Remark}{Remark}

\usepackage[pagebackref,breaklinks,colorlinks]{hyperref}

\usepackage[capitalize]{cleveref}
\crefname{section}{Sec.}{Secs.}
\Crefname{section}{Section}{Sections}
\Crefname{table}{Table}{Tables}
\crefname{table}{Tab.}{Tabs.}

\def\cvprPaperID{2856} 
\def\confName{CVPR}
\def\confYear{2022}

\begin{document}

\title{Spectral Unsupervised Domain Adaptation for Visual Recognition}

\author{Jingyi Zhang\textsuperscript{\rm 1,2,}\thanks{Equal contribution} \ \ Jiaxing Huang\textsuperscript{\rm 2,}$^*$ \ \ Zichen Tian\textsuperscript{\rm 1} \ \ Shijian Lu\textsuperscript{\rm 1,2,}\thanks{Corresponding author} 
\\
$^1$ S-lab $^2$ School of Computer Science and Engineering, Nanyang Technological University
\\
{\tt\small \{Jingyi.Zhang, Jiaxing.Huang, Zichen.Tian, Shijian.Lu\}@ntu.edu.sg}
}
\maketitle

\begin{abstract}
Though unsupervised domain adaptation (UDA) has achieved very impressive progress recently, it remains a great challenge due to missing target annotations and the rich discrepancy between source and target distributions. We propose Spectral UDA (SUDA), an effective and efficient  UDA technique that works in the spectral space and can generalize across different visual recognition tasks. SUDA addresses the UDA challenges from two perspectives. First, it introduces a \textit{spectrum transformer} (ST) that mitigates inter-domain discrepancies by enhancing domain-invariant spectra while suppressing domain-variant spectra of source and target samples simultaneously. Second, it introduces \textit{multi-view spectral learning} that learns useful unsupervised representations by maximizing mutual information among multiple ST-generated spectral views of each target sample. Extensive experiments show that SUDA achieves superior accuracy consistently across different visual tasks in object detection, semantic segmentation and image classification. 
Additionally, SUDA also works with the transformer-based network and achieves state-of-the-art performance on object detection.

\end{abstract}

\section{Introduction}
Deep learning techniques~\cite{krizhevsky2012alexnet,simonyan2014very, he2016resnet} have achieved great success in various visual recognition tasks such as image classification~\cite{krizhevsky2012alexnet,simonyan2014very,he2016resnet}, image segmentation~\cite{long2015fully,ronneberger2015u,badrinarayanan2017segnet, chen2017deeplab} and object detection~\cite{girshick2014rcnn, girshick2015fastrcnn, ren2015fasterrcnn, redmon2016yolo, liu2016ssd, carion2020detr}. The great success is at the price of large quantities of annotated training data which are often prohibitively laborious and time-consuming to collect~\cite{coco,deng2009imagenet,everingham2015pascal, cordts2016cityscapes}. One alternative that could mitigate this constraint is to leverage the off-the-shelf labeled data from one or multiple related \textit{source domains}. However, the model trained with source-domain data often experiences clear performance drop while applied to a \textit{target domain} where the data often have discrepant distributions as compared with the source-domain data~\cite{saito2018maximum,chen2018wild,tsai2018learning}.

\begin{figure}[t]
\centering
\includegraphics[width=1.0\linewidth]{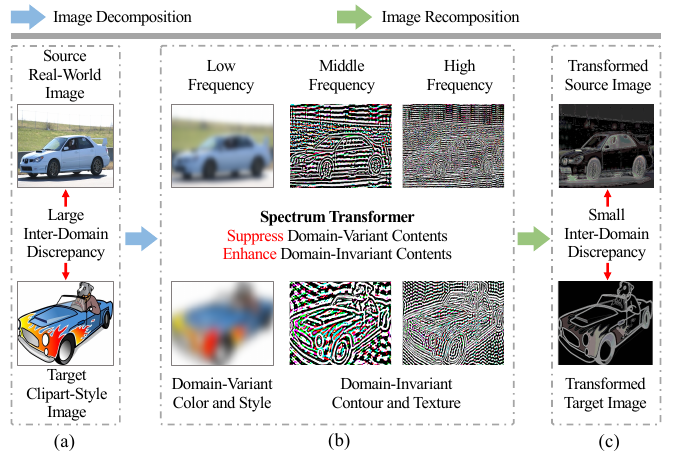}
\caption{
Illustration of the proposed spectrum transformer (ST): For images of different domains with clear distribution discrepancies as shown in (a), ST converts them into frequency space and decomposes the converted frequency signals into multiple frequency components (FCs) in low, middle, and high frequency bands as shown in (b).
It learns to identify and enhance domain-invariant FCs and suppress domain-variant FCs which effectively mitigates the inter-domain discrepancy as shown in (c). 
Note we increase the image contrast for better visualizing (c).
}
\label{fig:intro}
\end{figure}

Unsupervised domain adaptation (UDA) has been explored to mitigate the discrepancy between source and target domains. One typical approach is image-to-image translation with generative adversarial networks (GANs) which aligns source and target data in the input space by modifying source data to have similar styles as target data~\cite{inoue2018weakly, kim2019diversify, li2019bidirectional_seg}. However, image-to-image translation needs to learn large amounts of parameters which is usually computationally intensive. In addition, it impairs the end-to-end feature of UDA as it needs to train GANs first before applying them for image translation~\cite{inoue2018weakly, kim2019diversify, li2019bidirectional_seg}. Further, it could degrade UDA by undesirably modifying domain-invariant image structures that are closely entangled with domain-variant image styles in the spatial space~\cite{goodfellow2014gan,zhu2017cycle-gan}.

We propose Spectral UDA (SUDA) that tackles UDA challenges by learning domain-invariant spectral features efficiently and effectively. SUDA works from two perspectives. First, it introduces a spectrum transformer (ST) that learns to reduce inter-domain discrepancies by enhancing domain-invariant frequency components (FCs) and suppressing domain-variant FCs as illustrated in Fig.~\ref{fig:intro}. To this end, we design novel adversarial spectrum attention (ASA) that can identify domain-variant and domain-invariant FCs accurately.
Second, we design multi-view spectral learning (MSL) that learns diverse target representations by maximizing the mutual information among multiple ST-generated spectral views for each target sample. MSL introduces certain self-supervision which mitigates the lack of target annotations effectively.

The proposed SUDA has three desirable features. First, it is generalizable and performs consistently well across different visual tasks such as image classification, image segmentation and object detection. Second, it is an online and learnable technique whereas GANs-based image translation is offline and traditional image preprocessing is mostly non-learnable. Third, it is complementary with existing UDA methods and can be incorporated with consistent and clear performance boosts but little extra computation.

The contributions of this work are threefold.
\textit{First}, we designed SUDA that tackles UDA challenges effectively by learning domain-invariant spectral features.
\textit{Second}, we design an online learnable spectrum transformer that mitigates inter-domain discrepancy by enhancing domain-invariant FCs and suppressing domain-variant FCs simultaneously. To this end, we design ASA that leverages contextual information to identify domain-variant and domain-invariant FCs accurately. \textit{Third}, we design MSL that can learn diverse target representations by maximizing mutual information among multiple spectral views of each target sample. MSL mitigates the lack of target annotations effectively.

\begin{figure*}[ht]
\centering
\includegraphics[width=1.0\linewidth]{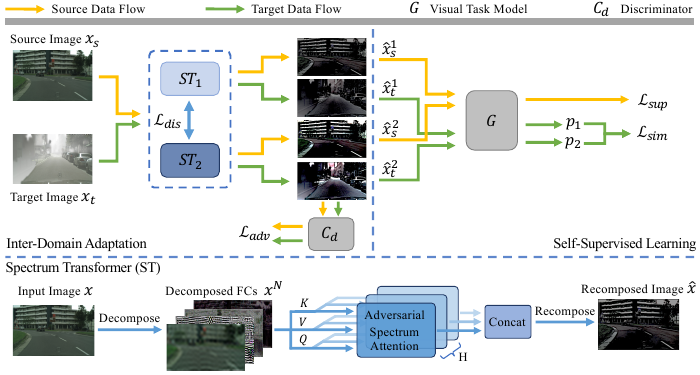}
\caption{
The overview of the proposed SUDA: For source image $x_s$ and target image $x_t$, SUDA first creates two spectral views for each of them with $ST_1$ and $ST_2$ (having different parameters via a discrepancy loss $\mathcal{L}_{dis}$) and then feeds the spectral views to a discriminator $C_d$ for adversarial learning for inter-domain adaptation. The ST outputs $\hat{x}^1_s$ and $\hat{x}^2_s$ are fed to a visual task model $G$ for supervised learning, while $\hat{x}^1_t$ and $\hat{x}^2_t$ are fed to $G$ for self-supervised learning under an unsupervised similarity loss $\mathcal{L}_{sim}$. The graph at the bottom shows more details of the ST design.
For an input image $x$, ST first transforms it to spectral representations which are then decomposed to $N$ FCs $x^N$. ASA then learns to identify and enhance domain-invariant FCs and suppress domain-variant FCs adaptively. Finally, the re-weighted FCs are recomposed back to a spatial-space image $\hat{x}$ for the ensuing supervised and self-supervised learning.
}
\label{fig:architecture}
\end{figure*}


\section{Related Works}
\textbf{Unsupervised Domain Adaptation.}
UDA has been studied extensively in recent years, largely for alleviating data annotation constraint in deep network training in various visual recognition tasks ~\cite{ganin2015grl, chen2018wild, zou2018self_seg, vu2019advent,luo2021unsupervised,huang2022multi,huang2021model,guan2021domain,guan2021uncertainty,huang2021cross,huang2020contextual,guan2021scale}. Besides adversarial learning~\cite{chen2018wild,tsai2018learning,he2019MAF,li2020SAP,2020coarse2fine,saito2019strong, vs2021mega} and self-training~\cite{yu2019self-training,kim2019self,zou2018self_seg,zou2019confidence}, image-to-image translation~\cite{li2019bidirectional_seg, kim2019diversify,inoue2018weakly} has been studied for reducing inter-domain discrepancy in the input space. To this end, a number of GANs~\cite{goodfellow2014generative, zhu2017cycle-gan} have been designed for translating image styles yet with minimal modification of image structures.
However, GAN training is usually time-consuming, which generally makes UDA frameworks not end-to-end trainable as they need to train GANs first before applying them to UDA.
In addition, GAN-based translation works in spatial space where image styles and image structures are closely entangled which inevitably modifies image structures undesirably. Some work~\cite{yang2020fda} attempts to translate images in frequency space by swapping certain pre-defined FCs of source and target images, but it is non-learnable and cannot accommodate individual images that usually have different spectral characteristics.
Recently, a few studies handle UDA via contrastive learning~\cite{huang2021model, zhang2021rpn} and adversarial attacking~\cite{huang2021rda}.

We design a spectrum transformer that learns to identify domain-variant and domain-invariant FCs for each individual image. It mitigates the inter-domain discrepancy by adaptively enhancing domain-invariant FCs and suppressing domain-variant FCs across source and target images.

\textbf{Learning in Frequency Space.} 
Image preprocessing in spectral space has been widely studied with various spectral filters in the traditional image processing studies~\cite{bow2002pattern,chaki2018beginner}. However, most traditional spectral preprocessing techniques are deterministic which handle each individual image in the same manner. Spectrum learning has attracted increasing attention recently with the advance of deep learning, and it has been studied for different vision tasks such as image translation~\cite{xu2020learningfrequency,durall2020watch,cai2021frequency}, image compression~\cite{xu2020learningfrequency},
network generalization~\cite{huang2021fsdr,xu2021fourier} and domain adaptation~\cite{yang2020phase,huang2021rda}.

We explored spectrum learning for the task of UDA. Specifically, we design multi-view spectral learning that generates different spectral views for each target image and maximizes their mutual information to learn diverse target representations without any image labels or annotations.

\textbf{Visual Attention.}
Visual attention has been widely studied in various visual recognition tasks. It can be broadly categorized into channel attention~\cite{hu2018squeezenet, qilong2020channelatt} and spatial attention~\cite{wang2018sptailatt} that aim to identify informative channels and spatial dependencies within each single channel, respectively.
Certain hybrid attention~\cite{woo2018cbam,cao2019gcnet,li2019sge} which combines channel and spatial attention has also been developed for better focus on informative image regions.
Recently, self-attention~\cite{vaswani2017transformer} has attracted increasing interest due to its powerful capability in learning spatial dependencies within input images. In addition, self-attention, which aggregates information across multiple self-attention at different image positions, has been explored in different visual recognition tasks~\cite{carion2020detr,zhu2020deformable,dosovitskiy2020ViT, zheng2020setr, xie2021segformer}.

We design ASA that introduces adversarial learning to help identify domain-variant and domain-invariant FCs. ASA works with multiple disentangled FCs which allows to model attention  effectively.

\textbf{Other Related Works.} Our proposed multi-view spectral learning is also related to consistency training~\cite{lai2021semi,ouali2020semi,chen2021semi,melas2021pixmatch,araslanov2021self}, which enforces prediction consistency between different views of an input image. For example, PixMatch~\cite{melas2021pixmatch} directly enforces pixel-level consistency of predictions from different data augmentations ($e.g$, CutMix, Fourier augmentation). SAC~\cite{araslanov2021self} acquires robust pseudo labels by fusing multiple predictions from different scales and retrains the network with data augmentations.

\section{Method}
\subsection{Task Definition}

This work focuses on UDA in different visual recognition tasks such as image classification, image segmentation, and object detection. It involves a labeled source domain $\mathcal{D}_s=\left\{\left(x_{s}^{{i}}, y_{s}^{{i}}\right)\right\}_{i=1}^{N_{{s}}}$, where $y_{s}^{{i}}$ is the label of the sample $x_{s}^{{i}}$, and an unlabeled target domain $\mathcal{D}_t = \left\{\left(x_{t}^{{i}}\right)\right\}_{i=1}^{N_{{t}}}$. The goal is to train a model $G$ that well performs in $\mathcal{D}_t$. The baseline model is trained with the data in $\mathcal{D}_s$ only:
\begin{equation}
\mathcal{L}_{sup} = l({G}(x_{s}), y_{s}), 
\label{baseline}
\end{equation}
where $l(\cdot)$ denotes a task-related loss, $e.g.$, the standard cross-entropy loss for image classification.

\subsection{Spectral Unsupervised Domain Adaptation}

We propose SUDA, an innovative spectral-space UDA technique that handles UDA by learning domain-invariant spectral features. SUDA has two key designs including a spectrum transformer for inter-domain adaptation and multi-view spectral learning for self-supervised learning. 

\textbf{Overview.} 
Fig.~\ref{fig:architecture} shows the framework of the proposed SUDA and the design of ST. Given a source-domain image $x_s\in\mathcal{D}_s$ and a target-domain image $x_t\in\mathcal{D}_t$, two complementary STs $ST_1$ and $ST_2$ (having different parameters via a discrepancy loss $\mathcal{L}_{dis}$) first transform the two images into spectral space and decompose them into multiple frequency components (FCs). The proposed ASA within ST then learns to identify and enhance domain-invariant FCs and suppress domain-variant FCs simultaneously via an adversarial loss $\mathcal{L}_{adv}$ with the discriminator $C_d$. ASA thus mitigates inter-domain discrepancies and leads to inter-domain adaptation effectively.

Here $\hat{x}_t^1$ and $\hat{x}_t^2$ as the output of $ST_1$ and $ST_2$ capture different spectral views of the target sample $x_t$.
They are fed to a visual task model ${G}$ for self-supervised learning, where the proposed MSL strives to maximize the mutual information among the two augmentations of $x_t$. Note we employ two STs for producing different spectral views of $x_t$ which allows to learn more diverse target representations.
SUDA can work with a single ST without MSL as well, more details to be discussed in the experiment section.

\begin{algorithm}[t]
    \caption{The proposed SUDA.
    }\label{algorithm_SUDA}
    \begin{algorithmic}[1]
        \REQUIRE Source domain $\mathcal{D}_s$; Target domain $\mathcal{D}_t$; Visual task model ${G}$; Spectrum transformers $ST_1$ and $ST_2$
        \ENSURE Learnt networks $ST_1$, $ST_2$ and $G$
        \FOR{$iter = 1$ \textbf{to} $Max\_Iter$}
            \STATE Sample a source data $\{x_s, y_s\} \in \mathcal{D}_s$ and a target data $x_t \in \mathcal{D}_t$
            \STATE \textbf{Inter-domain Adaptation:}
            \STATE \quad Calculate $ST_1(x_s)$, $ST_2(x_s)$, $ST_1(x_t)$ and $ST_2(x_t)$ by Eq.~\ref{multi-h}
            \STATE \quad Calculate $\mathcal{L}_{adv}$ by Eq.~\ref{l_adv}
            
            \STATE \textbf{Self-supervised Learning:}
            \STATE \quad  Calculate $\mathcal{L}_{dis}$ by Eq.~\ref{L_dis}
            \STATE \quad  Calculate $\mathcal{L}_{sim}$ by Eq.~\ref{L_sim}
            
            \STATE \textbf{Supervised Learning:}
            \STATE \quad  Calculate $\mathcal{L}_{sup}$ by Eq.~\ref{baseline}
            
            \STATE Optimize $ST_1$, $ST_2$ and visual task model $G$ by Eq.~\ref{all_loss}
        \ENDFOR
  \RETURN $ST_1$, $ST_2$ and $G$
  \end{algorithmic}
\end{algorithm}

\textbf{Spectrum Transformer:}
We design spectrum transformer for inter-domain adaptation. Given an image 
$x\in\mathbb{R}^{3 \times H \times W}$, 
ST first transforms it into spectral representation with Fast Fourier Transform. It then decomposes the 
spectral representation into $N$ FCs  
($i.e.$, $x^{N}=\left\{ x^n\right\}_{n=1}^{N}$ ,${x^n} \in \mathbb{R}^{3 \times 1 \times H \times W}$) 
evenly by using a band pass filter. 
The decomposed $x^N$ is fed to ASA within ST to enhance domain-invariant FCs and suppress domain-variant FCs adaptively. The design of ASA is shown at the bottom of Fig.~\ref{fig:architecture} and its definition is presented below.

\begin{Definition}
\label{attention}
The proposed ASA
is defined by:
\begin{equation}
\label{multi-h}
\begin{split}
ASA(x^N) = &\mathcal{W}_{cs}(\text{Concat} (A_1(x^N_*), A_2(x^N_*),...,\\&A_h(x^N_*),..., A_H(x^N_*)) \textbf{P}_H) \ x^N
\end{split}
\end{equation}
where $x^N_* \in \mathbb{R} ^ {3N \times 1}$ is pooled vector of $x^N$, $A_h(\cdot)$ is single attention head and $Concat(\cdot)$ denotes the concatenation of the outputs of $A_h(\cdot)$. $\textbf{P}_H \in \mathbb{R}^{H d_h \times d}$ projects the concatenation, where $d = 3N$ and $d_h = d / H$. Channel and spatial-wise attention $\mathcal{W}_{cs}(\cdot)$ takes the projected concatenation as input and further weights the decomposed $x^N$. Each single attention head $A_h(\cdot)$ is defined as a standard scaled dot-product attention, which maps a query ($Q$) and a set of key-value ($K$, $V$) pairs into an output:
\begin{equation}
\label{single_att_1}
(K, V, Q) = x^N_* \textbf{P}_{kvq},
\end{equation}
\begin{equation}
\label{single_att_2}
A_h(x^N_*) = \text{Softmax}(QK^T / \sqrt{d_h}) V,
\end{equation}
where the values of $K$, $V$ and $Q$ for each head are projected from input $x^N_*$ by $\textbf{P}_{kvq} \in \mathbb{R}^{d \times 3d_h}$.

\end{Definition}

In final, the output of the ASA 
is reshaped back to the size of $3 \times N \times H \times W$ and further recomposed to a full-spectrum spatial image $\hat{x} \in \mathbb{R}^{3 \times H \times W}$ by concatenation.
The output $\hat{x}$ from ST and its corresponding domain label (0 or 1) are then forwarded to discriminator $C_d$ for reducing inter-domain discrepancies in the input space. 
$C_d$ performs adversarial learning with an adversarial loss $\mathcal{L}_{adv}$:
\begin{equation}
\begin{split}
\label{l_adv}
\mathcal{L}_{adv} &= \mathbb{E} [\log {C_d}({ST}(x_s))] \\&+ \mathbb{E} [\log (1- {C_d}({ST}(x_t)))]
\end{split}
\end{equation}

\begin{Remark}
\label{rm-on-ST}
Note ST performs inter-domain adaptation by employing attention mechanisms which are essentially simple matrix multiplication operations. Compared with GANs that perform image translation, ST is much more efficient as it can be trained with visual task model $G$ in an end-to-end manner. Specifically, ST only involves $1$ attention layer with about $37,000$ parameters whereas a typical image translation GAN involves $9$ convolutional layers with about $11,000,000$ parameters
\cite{zhu2017cycle-gan}.
Due to the high computation costs of GANs, most GAN-based UDA methods~\cite{inoue2018weakly,kim2019diversify, arruda2019daynight} first train image translation GANs separately which ruins the end-to-end property of UDA undesirably.

\end{Remark}

\textbf{Multi-View Spectral Learning:} 
We develop multi-view spectral learning that exploits self-supervision for learning unsupervised target representation. For each target image $x_t$, SUDA creates two complementary spectral views $\hat{x}^1_t$ and $\hat{x}^2_t$ by employing two spectrum transformers $ST_1$ and $ST_2$. We enforce $ST_1$ and $ST_2$ to have different parameters by a discrepancy weight loss $\mathcal{L}_{dis}$ so that $ST_1$ and $ST_2$ can learn complementary domain-invariant FCs of $x_t$:
\begin{equation}
\label{L_dis}
\mathcal{L}_{dis}=\frac{\vec{\theta}_{1} \cdot \vec{\theta}_{2}}{\left\|\vec{\theta}_{1}\right\|\left\|\vec{\theta}_{2}\right\|},
\end{equation}
where $\vec{\theta}_1$ and $\vec{\theta}_2$ denote the parameters of $ST_1$ and $ST_2$.

The two complementary spectral views of $x_t$ are then forwarded to a visual task model $G$ which produces predictions $p_1 = G (\hat{x}^1_t)$ and $p_2 = G (\hat{x}^2_t)$. 
To maximize the mutual information~\cite{grill2020byol, chen2021siamese} of the two spectral views, we minimize a similarity loss $\mathcal{L}_{sim}$ between $p_1$ and $p_2$ by:

\begin{equation}
\label{L_sim}
\mathcal{L}_{sim} = ||p_1-p_2||.
\end{equation}

\begin{Remark}
\label{rm-on-MSL}
Note we implement two complementary STs in SUDA for learning more diverse domain-invariant spectral information. We also introduce a discrepancy loss $\mathcal{L}_{dis}$ in Eq.~\ref{L_dis} that enhances this feature by forcing the two STs to learn different parameters. Hence, a single ST learns to minimize inter-domain discrepancy, while the two STs learn more diverse information from different spectral views for each training sample.

\end{Remark}

\renewcommand\arraystretch{1.0}
\begin{table*}[t]
\centering
\begin{footnotesize}
\begin{tabular}{p{3.5cm}|*{1}p{1.3cm}|*{8}{p{0.7cm}}*{1}{p{0.6cm}}}
 \toprule
 \multicolumn{11}{c}{\textbf{Cityscapes $\rightarrow$ Foggy cityscapes Object Detection}} \\
\hline
 Methods& Backbone &person & rider & car & truck & bus & train & mcycle & bicycle & mAP\\
\hline
Faster R-CNN~\cite{ren2015fasterrcnn} (Baseline) &ResNet-50 & 26.9 & 38.2 & 35.6 &  18.3 & 32.4 & 9.6 & 25.8 & 28.6 & 26.9 \\
\hline
DAF~\cite{chen2018wild} &ResNet-50 &29.2 & 40.4 & 43.4 & 19.7 & 38.3 & 28.5 & 23.7 & 32.7 & 32.0\\
\textbf{+SUDA} & ResNet-50&39.5 &46.8 &54.6 & 29.3 & 50.7 & 44.6 & 31.6 & 39.5 & \textbf{42.1}\\
\hline
SCDA~\cite{zhu2019selective} &ResNet-50& 33.8 & 42.1 & 52.1 & 26.8 & 42.5 & 26.5 & 29.2 & 34.5 & 35.9 \\
\textbf{+SUDA} &ResNet-50 & 39.7 &47.7 &54.3 &27.6 &51.8 &46.5 &31.2 &39.6 &\textbf{42.3}
 \\
\hline
SWDA~\cite{saito2019strong}&ResNet-50 & 31.8 & 44.3 & 48.9 & 21.0 & 43.8 & 28.0 & 28.9 & 35.8 & 35.3 \\
\textbf{+SUDA} & ResNet-50&39.5 &48.2 &57.8 &29.5 &52.9 &37.5 &34.5 &41.3 &\textbf{42.7}
 \\
\hline
\textbf{SUDA}&ResNet-50 &40.2 & 47.9 & 54.6 & 28.5 & 49.5 & 39.2 & 33.8 & 41.5 & \textbf{41.9}\\
\hline
\hline
DETR~\cite{zhu2020deformable} (Baseline) & ResNet-50 & 43.7 & 38.0 & 57.2 & 15.2 & 34.7 & 14.4 & 26.1 & 42.4 & 34.0 \\
\hline
DAF~\cite{chen2018wild} & ResNet-50 & 49.4 & 49.7 & 62.1 & 23.6 & 43.8 & 21.6 & 31.3 & 43.1
& 40.6
\\
\textbf{+SUDA}&ResNet-50 &50.5 &51.7 &64.1 &26.7 &48.5 &14.2 &38.1 &49.5 &\textbf{42.9}
\\
\hline
SWDA~\cite{saito2019strong} & ResNet-50 & 49.0 & 49.0 & 61.4 & 23.9 & 43.1 & 22.9 & 31.0 & 45.2 & 40.7
\\
\textbf{+SUDA}& ResNet-50 &50.7 &50.3 &67.3 &22.3 &45.2 &27.4 &34.0 &48.9 &\textbf{43.3}
\\
\hline
CRDA~\cite{xu2020category} & ResNet-50 & 49.8 & 48.4 & 61.9 & 22.3 & 40.7 & 30.0 & 29.9 & 45.4 & 41.1
\\
\textbf{+SUDA}& ResNet-50 & 52.3 & 51.6 & 66.7 & 30.4 & 47.1 & 11.9 & 36.8 & 48.7 & \textbf{43.2}
\\
\hline
CF~\cite{2020coarse2fine} & ResNet-50 & 49.6 & 49.7 & 62.6 & 23.3 & 43.4 & 27.4 & 30.2 & 44.8 & 41.4
\\
\textbf{+SUDA}& ResNet-50 & 51.2 & 51.4 & 68.5& 25.3& 48.0& 26.5& 33.8& 49.9& \textbf{44.3}
\\
\hline
SAP~\cite{li2020SAP} & ResNet-50 & 49.3 & 49.9 & 62.5 & 23.0 & 44.1 & 29.4 & 31.3 & 45.8 & 41.9
\\
\textbf{+SUDA}&ResNet-50 & 51.4 & 52.2 & 67.5 & 28.7 & 49.6 & 28.7& 39.2& 50.4& \textbf{46.0}
\\
\hline
\textbf{SUDA} & ResNet-50 &50.5 & 51.7 & 64.1 & 26.7 & 48.5 & 13.1 & 38.1 & 49.5 & \textbf{42.8}\\
\bottomrule
\end{tabular}
\end{footnotesize}
\caption{Experiments on UDA-based object detection task Cityscapes $\rightarrow$ Foggy Cityscapes.}
\label{table:det_city2fog}
\end{table*}

\renewcommand\arraystretch{1.0}
\begin{table*}[t]
\centering
\begin{footnotesize}
\begin{tabular}{p{1.3cm}|*{20}{p{0.32cm}}*{1}{p{0.31cm}}}
\toprule
 \multicolumn{22}{c}{\textbf{PASCAL VOC $\rightarrow$ Clipart1k Object Detection}} \\
 \hline
Methods & aero & bcyc. & bird & boat & bott. & bus & car & cat & chair & cow & table & dog & horse & bike & pers. & plant & sheep & sofa & train & tv& mAP\\
\hline
Baseline~\cite{ren2015fasterrcnn} & 35.6 & 52.5 & 24.3 & 23.0 & 20.0 & 43.9 & 32.8 & 10.7 & 30.6 & 11.7 & 13.8 & 6.0 & 36.8 & 45.9 & 48.7 & 41.9 & 16.5 & 7.3 & 22.9 & 32.0 & 27.8\\
\hline
DAF~\cite{chen2018wild} & 15.0 & 34.6 & 12.4 & 11.9 & 19.8 & 21.1 & 23.2 & 3.1 & 22.1 & 26.3 & 10.6 & 10.0 & 19.6 & 39.4 & 34.6 & 29.3 & 1.0 & 17.1 & 19.7 & 24.8 & 19.8 \\
\textbf{+SUDA} & 28.2 & 53.8 & 37.1 & 15.4 & 37.6 & 66.6 & 35.3 & 21.7 & 38.7 & 48.7 & 18.3 & 28.4 & 24.4 & 82.4 & 61.0 & 44.5 & 11.9 & 34.4 & 49.5 & 59.7 & \textbf{39.9}\\
\hline
SWDA~\cite{saito2019strong}&26.2 & 48.5 & 32.6 & 33.7 & 38.5 & 54.3 & 37.1 & 18.6 & 34.8 & 58.3 & 17.0 & 12.5 & 33.8 & 65.5 & 61.6 & 52.0 & 9.3 & 24.9 & 54.1 & 49.1 & 38.1\\
\textbf{+SUDA} &33.7 & 61.8 & 36.9 & 23.1 & 39.2 & 56.2 & 33.9 & 23.4 & 38.7 & 45.9 & 15.4 & 23.4 & 25.8 & 75.8 & 58.6 & 41.8 & 15.7 & 33.2 & 61.7 & 60.1 & \textbf{40.2}\\
\hline
\textbf{SUDA} & 33.8 & 56.1 & 32.3 & 24.1 & 30.9 & 54.6 & 38.5 & 18.4 & 34.0 & 41.0 & 18.7 & 24.3 & 29.5 & 84.0 & 57.6 & 49.1 & 14.3 & 36.3 & 55.6 & 51.6 & \textbf{39.2}\\
\bottomrule
\end{tabular}
\end{footnotesize}
\caption{Experiments on UDA-based object detection task PASCAL VOC $\rightarrow$ Clipart1k.}
\label{table:voc2cli}
\end{table*}

\renewcommand\arraystretch{1.0}
\begin{table}[t]
\centering
\begin{footnotesize}
\begin{tabular}{p{3.1cm}|*{4}{p{0.5cm}}|*{1}{p{0.5cm}}}
\toprule
 \multicolumn{6}{c}{\textbf{Cityscapes $\rightarrow$ Foggy cityscapes}} \\
 \hline
Methods &$L_{sup}$ &$L_{adv}$ &$L_{dis}$ &$L_{sim}$ &mAP \\
\hline
Baseline~\cite{zhu2020deformable} &\checkmark &  &  &  &34.0  \\
+Single ST &\checkmark &\checkmark  &  &  &40.6   \\ 
+Two STs &\checkmark &\checkmark  &\checkmark &  &41.8   \\
+Two STs +MSL\textbf{(SUDA)} &\checkmark &\checkmark &\checkmark  &\checkmark  &\textbf{42.8}  \\
\bottomrule
\end{tabular}
\end{footnotesize}
\caption{Ablation study of the proposed Spectrum Transformer and Multi-view Spectral Learning over object detection task Cityscapes $\rightarrow$ Foggy Cityscapes.}
\label{table:ablation_city2fog}
\end{table}

\textbf{Overall Training Objective.} 
The objective of SUDA consists of three losses as stated in Algorithm~\ref{algorithm_SUDA}, namely, the supervised task loss $\mathcal{L}_{sup}$ in Eq.~\ref{baseline}, the inter-domain adaptation loss $\mathcal{L}_{adv}$ in Eq.~\ref{l_adv}, and the self-supervised learning loss $\mathcal{L}_{self}$ which consists of the discrepancy loss $\mathcal{L}_{dis}$ and the similarity loss $\mathcal{L}_{sim}$ in Eqs.~\ref{L_dis} and \ref{L_sim}. The overall training objective can thus be formulated by
\begin{equation}
\label{all_loss}
\max _{C_d} \min _{{G}, {ST}} \mathcal{L}_{sup}-\lambda_{c} \mathcal{L}_{adv} + \lambda_{s}\mathcal{L}_{self},
\end{equation}
where $\lambda_{c}$ and $\lambda_{s}$ denote the balance weights.

\renewcommand\arraystretch{1.0}
\begin{table*}[t]
\centering
\begin{footnotesize}
\begin{tabular}{p{2.4cm}|*{12}{p{0.7cm}}*{1}{p{0.55cm}}}
\toprule
 \multicolumn{14}{c}{\textbf{VisDA17 Classification}} \\
 \hline
Methods & aero. & bike & bus & car & horse & knife & motor & person & plant & skate. & train & truck & Mean\\
\hline
Res-101 \cite{he2016resnet} & 55.1 & 53.3 & 61.9 & 59.1 & 80.6 & 17.9 & 79.7 & 31.2 & 81.0 & 26.5 & 73.5 & 8.5 & 52.4\\
MCD \cite{saito2018maximum} & 87.0 & 60.9 & {83.7} & 64.0 & 88.9 & 79.6 & 84.7 & {76.9} & {88.6} & 40.3 & 83.0 & 25.8 & 71.9\\
ADR \cite{saito2018adversarial} & 87.8 & 79.5 & {83.7} & 65.3 & {92.3} & 61.8 & {88.9} & 73.2 & 87.8 & 60.0 & {85.5} & {32.3} & 74.8\\  
SimNet-Res152 \cite{pinheiro2018unsupervised} & {94.3} & 82.3 & 73.5 & 47.2 & 87.9 & 49.2 & 75.1 & 79.7 & 85.3 & 68.5 & 81.1 & 50.3 & 72.9\\
GTA-Res152 \cite{sankaranarayanan2018generate} & - & - & - & - & - & - & - & - & - & - & - & - & 77.1\\
\hline
CBST~\cite{zou2018self_seg} &87.2 & 78.8 & 56.5 & 55.4 & 85.1 & 79.2 & 83.8 &  77.7 & 82.8 & 88.8 & 69.0 & 72.0 & 76.4\\
\textbf{+SUDA} &89.6	&79.0	&69.0	&66.1	&88.5	&79.9	&86.7	&79.6	&85.4	&87.7	&81.0	&73.8	&\textbf{80.5}\\\hline
CRST~\cite{zou2019confidence} & 88.0 & 79.2 & 61.0 & 60.0 & 87.5 & 81.4 & 86.3 & 78.8 & 85.6 & 86.6 & 73.9 &   68.8 &78.1\\
\textbf{+SUDA} &91.5	&79.7	&71.9	&66.5	&88.5	&81.1	&85.6	&79.5	&86.2	&86.5	&79.9	&74.3	&\textbf{80.9}\\
\hline
\textbf{SUDA} &88.3	&79.3	&66.2	&64.7	&87.4	&80.1	&85.9	&78.3	&86.3	&87.5	&78.8	&74.5	&\textbf{79.8}\\
\bottomrule
\end{tabular}
\end{footnotesize}
\caption{Experiments on UDA-based image classification task  VisDA17.}
\label{table:visda17}
\end{table*}

\renewcommand\arraystretch{1.0}
\begin{table}[t]
\centering
\begin{footnotesize}
\begin{tabular}{p{1.75cm}|*{6}{p{0.42cm}}*{1}{p{0.43cm}}}
\toprule
 \multicolumn{8}{c}{\textbf{Office-31 Classification}} \\
 \hline
Methods & A$\rightarrow$W & D$\rightarrow$W & W$\rightarrow$D & A$\rightarrow$D & D$\rightarrow$A & W$\rightarrow$A & Mean\\
\hline
ResNet-50 \cite{he2016resnet} & 68.4 & 96.7 & 99.3 & 68.9 & 62.5 & 60.7 & 76.1\\
JAN \cite{long2017deep} & 85.4 & 97.4 & 99.8 & 84.7 & 68.6 & 70.0 & 84.3\\
GTA \cite{sankaranarayanan2018generate} & {89.5} & 97.9 & 99.8 & 87.7 & 72.8 & 71.4 & 86.5\\
\hline
CBST~\cite{zou2018self_seg} & 87.8 & 98.5 & {100.0} & 86.5 & 71.2 & 70.9 & 85.8\\
\textbf{+SUDA}   &90.5	&98.6	&100.0	&91.4	&72.7	&72.1	&\textbf{87.6}
\\\hline
CRST~\cite{zou2019confidence} & 89.4 & {98.9} & {100.0} & 88.7 & 72.6 & 70.9 & 86.8\\
\textbf{+SUDA} &91.0	&98.8	&100.0	&91.9	&72.9	&72.3	&\textbf{87.8}
\\\hline
\textbf{SUDA} &90.8	&98.7	&100.0	&91.2	&72.2	&71.4	&\textbf{87.4}\\
\bottomrule
\end{tabular}
\end{footnotesize}
\caption{Experiments on UDA-based image classification task Office-31.}
\label{table:office}
\end{table}

\section{Experiment}

This section presents experiments including datasets and implementation details, domain adaptation evaluations for object detection, image classification, and semantic segmentation tasks, and discussion, respectively. More details are to be described in the ensuing subsections.

\subsection{Datasets}
\label{datasets}
We evaluate SUDA over multiple datasets across different visual UDA tasks on object detection, image classification and semantic segmentation as listed:

\noindent\textbf{UDA for Object Detection:} 
We study two object detection tasks Cityscapes~\cite{cordts2016cityscapes} $\rightarrow$ Foggy Cityscapes~\cite{sakaridis2018foggy} and PASCAL VOC~\cite{everingham2015pascal} $\rightarrow$ Clipart1k~\cite{inoue2018weakly}. 

\noindent\textbf{UDA for Image Classification:} We study two UDA-based image classification tasks VisDA17~\cite{peng2018visda} and Office-31~\cite{saenko2010adapting}. For VisDA17, we evaluate the task synthetic$\rightarrow$real. For Office-31, we study six adaptation tasks: A$\rightarrow$W, D$\rightarrow$W, W$\rightarrow$D, A$\rightarrow$D, D$\rightarrow$A, and W$\rightarrow$A.

\noindent\textbf{UDA for Semantic Segmentation:} We study two synthetic-to-real semantic segmentation tasks GTA5 \cite{richter2016playing} $\rightarrow$ Cityscapes \cite{cordts2016cityscapes} and SYNTHIA \cite{ros2016synthia} $\rightarrow$ Cityscapes.

Due to the space limit, we provide more details about datasets in Section A.1 in supplementary materials.

\subsection{Implementation Details}
\label{implement}

\textbf{Object Detection:} For Cityscapes$\rightarrow$ Foggy Cityscapes, we adopt Faster R-CNN~\cite{ren2015fasterrcnn} and deformable-DETR~\cite{zhu2020deformable} as detection networks and ResNet-50~\cite{he2016resnet} as backbone as in~\cite{cai2019mtor, zhu2020deformable}.
For PASCAL VOC $\rightarrow$ Clipart1k, we adopt Faster R-CNN with ResNet-101~\cite{he2016resnet} as in~\cite{inoue2018weakly, saito2019strong}.

\textbf{Image Classification:} Following~\cite{zou2019confidence, saenko2010adapting}, we use ResNet-101 and ResNet-50~\cite{he2016resnet} as backbones for the tasks VisDA17 and Office-31, respectively.

\textbf{Semantic Segmentation:} We use DeepLab-V2~\cite{chen2017deeplab} with ResNet-101 \cite{he2016resnet} as the segmentation network as in~\cite{tsai2018learning,zou2018self_seg}.

For all visual recognition tasks, we set the number of FCs $N$ at $32$. Due to the space limit,  we provide more implementation details in Section A.2 in supplementary materials.

\renewcommand\arraystretch{1.0}
\begin{table*}[t]
\centering
\begin{footnotesize}
\begin{tabular}{p{1.5cm}|*{19}{p{0.32cm}}p{0.5cm}}
 \toprule
 \multicolumn{21}{c}{\textbf{GTA5 $\rightarrow$ Cityscapes Semantic Segmentation}} \\
 \hline
 Methods  & road & side. & buil. & wall & fence & pole & light & sign & vege. & ter. & sky & pers. & rider & car & truck & bus & train & mot. & bike & mIoU\\
 \hline
 Baseline~\cite{he2016resnet} &75.8	&16.8	&77.2	&12.5	&21.0	&25.5	&30.1	&20.1	&81.3	&24.6	&70.3	&53.8	&26.4	&49.9	&17.2	&25.9	&6.5	&25.3	&36.0	&36.6\\
AdaptSeg~\cite{tsai2018learning}  &86.5	&36.0	&79.9	&23.4	&23.3	&23.9	&35.2	&14.8	&83.4	&33.3	&75.6	&58.5	&27.6	&73.7	&32.5	&35.4	&3.9	&30.1	&28.1	&42.4\\
CBST~\cite{zou2018self_seg} &91.8 &53.5 &80.5 &32.7 &21.0 &34.0 &28.9 &20.4 &83.9 &34.2 &80.9 &53.1 &24.0 &82.7 &30.3 &35.9 &16.0 &25.9 &42.8 &45.9\\
AdvEnt~\cite{vu2019advent}  &89.4 &33.1 &81.0 &26.6 &26.8 &27.2 &33.5 &24.7 &{83.9} &{36.7} &78.8 &58.7 &30.5 &{84.8} &38.5 &44.5 &1.7 &31.6 &32.4 &45.5\\
CRST~\cite{zou2019confidence}  &91.0 &55.4 &80.0 &33.7 &21.4 &37.3 &32.9 &24.5 &85.0 &34.1 &80.8 &57.7 &24.6 &84.1 &27.8 &30.1 &26.9 &26.0 &42.3 &47.1\\
BDL~\cite{li2019bidirectional_seg}  &91.0  &44.7  &84.2  &34.6  &27.6 &30.2  &36.0  &36.0 &85.0  &43.6  &83.0 &58.6 &31.6  &83.3  &35.3  &49.7 &3.3  &28.8  &35.6 &48.5\\
        
CrCDA~\cite{huang2020contextual}  &92.4	&55.3	&82.3	&31.2	&29.1	&32.5	&33.2	&35.6	&83.5	&34.8	&84.2	&58.9	&32.2	&84.7	&40.6	&46.1	&2.1	&31.1	&32.7	&48.6 \\
\hline
RDA~\cite{huang2021rda} &89.8 &39.1 &81.7 &27.6 &19.9& 34.2& 35.9 &23.3& 82.1& 29.5& 76.6 &58.3 &26.0& 82.1& 32.5& 45.2 &15.3 &26.9 &33.5 &45.2\\
\textbf{+SUDA} &91.5	&52.1	&82.2	&32.3	&24.2	&36.2	&44.3	&36.3	&84.1	&39.4	&78.3	&59.6	&26.2	&83.7	&37.5	&45.8	&12.4	&27.7	&39.0	&\textbf{49.1}\\
\hline
TIR~\cite{kim2020learning}  &92.9 &55.0 &85.3 &34.2 &31.1 &34.9 &40.7 &34.0 &85.2 &40.1 &87.1 &61.0 &31.1 &82.5 &32.3 &42.9 &0.3 &36.4 &46.1 &50.2\\
\textbf{+SUDA}  &92.6	&54.9	&85.9	&31	&30.6	&37.6	&43.6	&41.3	&84.5	&39.3	&87	&60.4	&32.6	&84.6	&38.3	&46.7	&11.2	&34.9	&43.7	&\textbf{51.6}\\
\hline
FDA~\cite{yang2020fda} & 92.5 &53.3 &82.4 &26.5 &27.6 &36.4 &40.6 &38.9 &82.3 &39.8 &78.0 &62.6 &34.4 &84.9 &34.1 &53.1 &16.9 &27.7 &46.4 &50.5\\
\textbf{+SUDA}  &93.4	&55.1	&84.9	&31.5	&28.9	&38.3	&45.6	&41.9	&84.6	&40.0	&83.1	&61.4	&31.3	&84.8	&41.1	&50.5	&15.4	&30.8	&43.9	&\textbf{51.9}\\
\hline
ProDA~\cite{zhang2021proda} &87.8& 56.0& 79.7 &46.3& 44.8& 45.6& 53.5& 53.5& 88.6& 45.2& 82.1& 70.7& 39.2& 88.8& 45.5& 59.4& 1.0& 48.9& 56.4& 57.5\\
\textbf{+SUDA} & 94.5& 67.5& 86.4& 45.1& 41.4& 47.1& 50.5& 55.6& 89.6& 48.1& 87.4& 67.3& 1.1& 88.9& 39.1& 60.2& 33.3& 44.5& 61.1& \textbf{58.3}\\
\hline
\textbf{SUDA}  &91.1	&52.3	&82.9	&30.1	&25.7	&38.0	&44.9	&38.2	&83.9	&39.1	&79.2	&58.4	&26.4	&84.5	&37.7	&45.6	&10.1	&23.1	&36.0	&\textbf{48.8}\\
\bottomrule
\end{tabular}
\end{footnotesize}
\caption{Experiments on UDA-based semantic segmentation task GTA5 $\rightarrow$ Cityscapes.}
\label{table:gta2city}
\end{table*}

\renewcommand\arraystretch{1.0}
\begin{table*}[t]
\centering
\begin{footnotesize}
\begin{tabular}{p{2cm}|*{16}{p{0.35cm}}p{0.55cm}p{0.75cm}}
 \toprule
 \multicolumn{19}{c}{\textbf{SYNTHIA $\rightarrow$ Cityscapes Semantic Segmentation}} \\
 \hline
 Methods  &{road} &{side.} &{buil.}&{wall} &{fence} &{pole} &{light} &{sign} &{vege.} &{sky} &{pers.} &{rider} &{car} &{bus} &{mot.} &{bike} &mIoU  &mIoU*\\
 \hline
 Baseline~\cite{he2016resnet} &55.6	&23.8	&74.6	&9.2	&0.2	&24.4	&6.1	&12.1	&74.8	&79.0	&55.3	&19.1	&39.6	&23.3	&13.7	&25.0	&33.5	&38.6\\
AdaptSeg~\cite{tsai2018learning}  &84.3 &42.7 &77.5 &- &- &- &4.7 &7.0 &77.9 &82.5 &54.3 &21.0 &72.3 &32.2 &18.9 &32.3 &- &46.7\\
AdvEnt~\cite{vu2019advent}  &85.6 &42.2 &79.7 &{8.7} &0.4 &25.9 &5.4 &8.1 &{80.4} &84.1 &{57.9} &23.8 &73.3 &36.4 &14.2 &{33.0} &41.2 &48.0\\
CrCDA~\cite{huang2020contextual}   &86.2	&44.9	&79.5	&8.3	&0.7	&27.8	&9.4	&11.8	&78.6	&86.5	&57.2	&26.1	&76.8	&39.9	&21.5	&32.1	&42.9	&50.0\\
CRST~\cite{zou2019confidence}  &67.7 &32.2 &73.9 &10.7 &1.6 &37.4 &22.2 &31.2 &80.8 &80.5 &60.8 &29.1 &82.8 &25.0 &19.4 &45.3 &43.8 &50.1\\
\hline
TIR~\cite{kim2020learning}  &92.6 &53.2 &79.2 &- &- &- &1.6 &7.5 &78.6 &84.4 &52.6 &20.0 &82.1 &34.8 &14.6 &39.4 &- &49.3\\
\textbf{+SUDA} &83.9	&40.1	&76.9	&4.5	&0.1	&26.1	&22.9	&26.4	&79.6	&80.7	&58.1	&28.3	&81.0	&37.4	&35.1	&46.8	&45.5	&\textbf{53.6}\\\hline
FDA~\cite{yang2020fda}  &79.3 &35.0 &73.2 &- &- &- &19.9 &24.0 &61.7 &82.6 &61.4 &31.1 &83.9 &40.8 &38.4 &51.1 &- &52.5\\
\textbf{+SUDA} &85.6	&38.8	&76.7	&9.2	&0.2	&28.4	&25.4	&27.0	&78.4	&81.7	&60.4	&28.6	&82.8	&38.8	&36.2	&48.1	&46.7	&\textbf{54.5}\\\hline
\textbf{SUDA} &83.4	&36.0	&71.3	&8.7	&0.1	&26.0	&18.2	&26.7	&72.4	&80.2	&58.4	&30.8	&80.6	&38.7	&36.1	&46.1	&44.6	&\textbf{52.2}\\
\bottomrule
\end{tabular}
\end{footnotesize}
\caption{Experiments on UDA-based semantic segmentation task SYNTHIA $\rightarrow$ Cityscapes. mIoU is evaluated on 16 classes, and mIoU* is evaluated on 13 classes.}
\label{table:synthia2city}
\end{table*}

\subsection{Domain Adaptive Object Detection}
\label{detection}

We first benchmark the proposed SUDA with state-of-the-art domain adaptive object detection methods over two UDA tasks Cityscapes$\rightarrow$ Foggy Cityscapes and PASCAL VOC $\rightarrow$ Clipart1k. Tables~\ref{table:det_city2fog} and~\ref{table:voc2cli} show experimental results. It can be seen that SUDA achieves competitive object detection performance as compared with all highly-optimized state-of-the-art methods across two very different network architectures (Faster R-CNN and deformable-DETR). In addition, SUDA is complementary to most existing methods which produces clear and consistent performance boosts while incorporated as a plug-in.

We also examine the proposed SUDA by performing several ablation studies over a domain adaptive object detection task Cityscapes$\rightarrow$ Foggy Cityscapes. Table~\ref{table:ablation_city2fog} shows experimental results. It can be seen that including either one or two STs in \textbf{+Single ST} and \textbf{+Two STs} outperforms the \textbf{Baseline} (deformable-DETR) by large margins. In addition, including two complementary STs in \textbf{+Two STs} perform clearly better than including a single ST in \textbf{+Single ST} as the two STs learn more diverse and complementary domain-invariant spectral information. Further including MSL beyond the two STs in \textbf{+Two STs +MSL (SUDA)} performs clearly the best, demonstrating the effectiveness of the proposed multi-view spectral learning.

\subsection{Domain Adaptive Image Classification}
\label{classification}

We evaluate and benchmark SUDA over two domain adaptive image classification tasks VisDA17 and Office-31. Tables~\ref{table:visda17} and~\ref{table:office} shows experimental results, where SUDA outperforms all state-of-the-art methods clearly. In addition, SUDA is complementary to existing methods which produces consistent and clear performance boosts while incorporated as a plug-in. Note we perform the complementary studies over a few representative domain adaptive image classification methods only due to space limit.

\subsection{Domain Adaptive Semantic Segmentation}
\label{segmentation}

We evaluate and benchmark SUDA over two domain adaptive semantic segmentation tasks GTA5 $\rightarrow$ Cityscapes and SYNTHIA $\rightarrow$ Cityscapes. Tables~\ref{table:gta2city} and~\ref{table:synthia2city} show experimental results. We can see that SUDA achieves competitive segmentation performance as compared with highly-optimized state-of-the-art methods. In addition, it is complementary with existing methods which produce consistent performance boosts while incorporated as a plug-in.

\section{Discussion}
\label{Discuss}

\textbf{Generalization across Visual Tasks:}
The proposed SUDA is generally applicable to various visual recognition tasks as described in Sections~\ref{detection},~\ref{classification} and~\ref{segmentation}.
With simple implementation and minimal fine-tuning as described in Section~\ref{implement}, it produces competitive performance consistently across different tasks as shown in Tables~\ref{table:det_city2fog}-\ref{table:synthia2city}. The superior generalization is largely attributed to the spectral transformer and multi-view spectral learning which are task-agnostic by learning domain-invariant spectra.

\begin{figure}
  \centering
  \begin{subfigure}{0.48\linewidth}
    \includegraphics[width=1.0\linewidth]{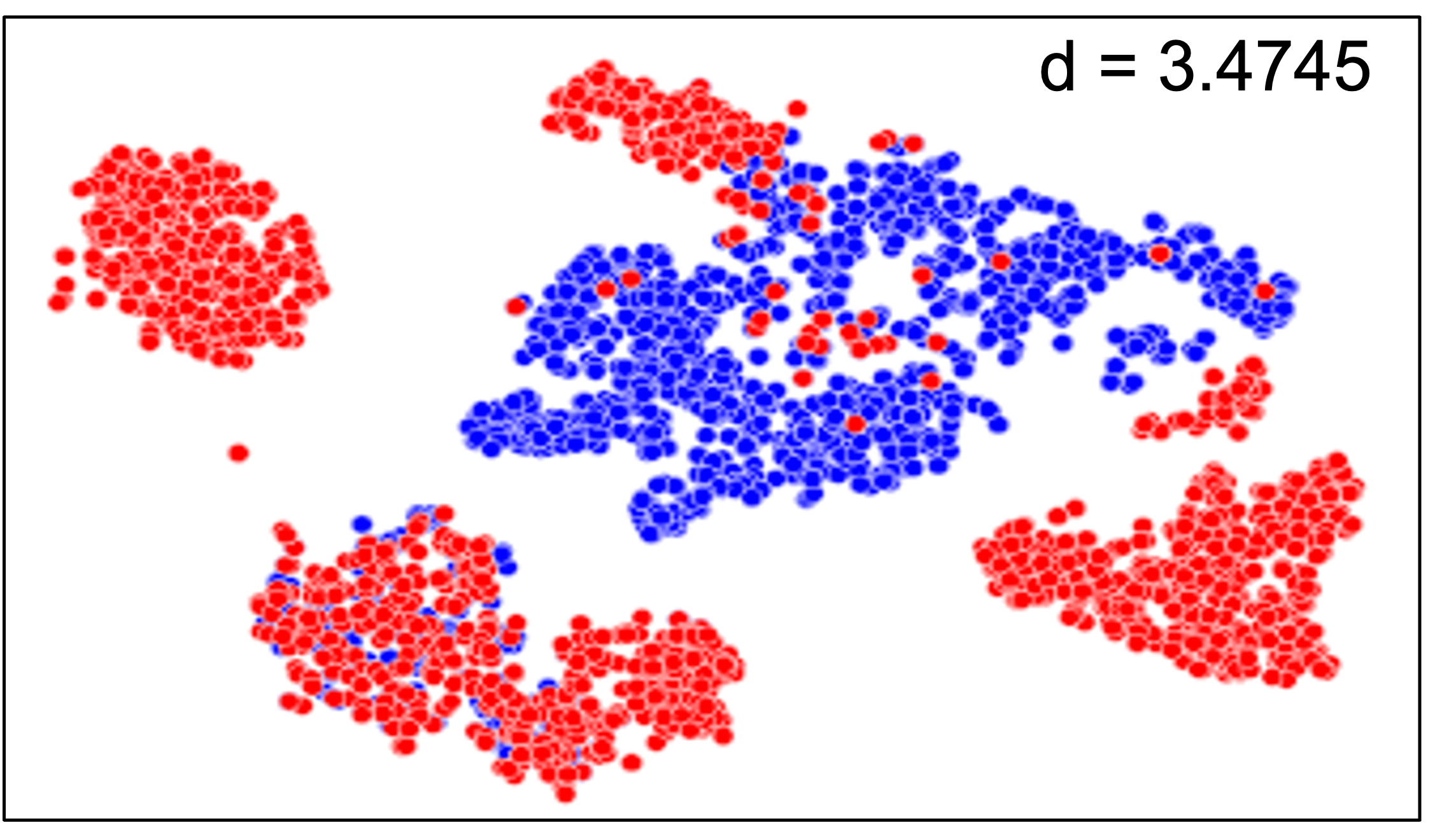}
    \caption{Original features}
  \end{subfigure}
  \hfill
  \begin{subfigure}{0.48\linewidth}
    \includegraphics[width=1.0\linewidth]{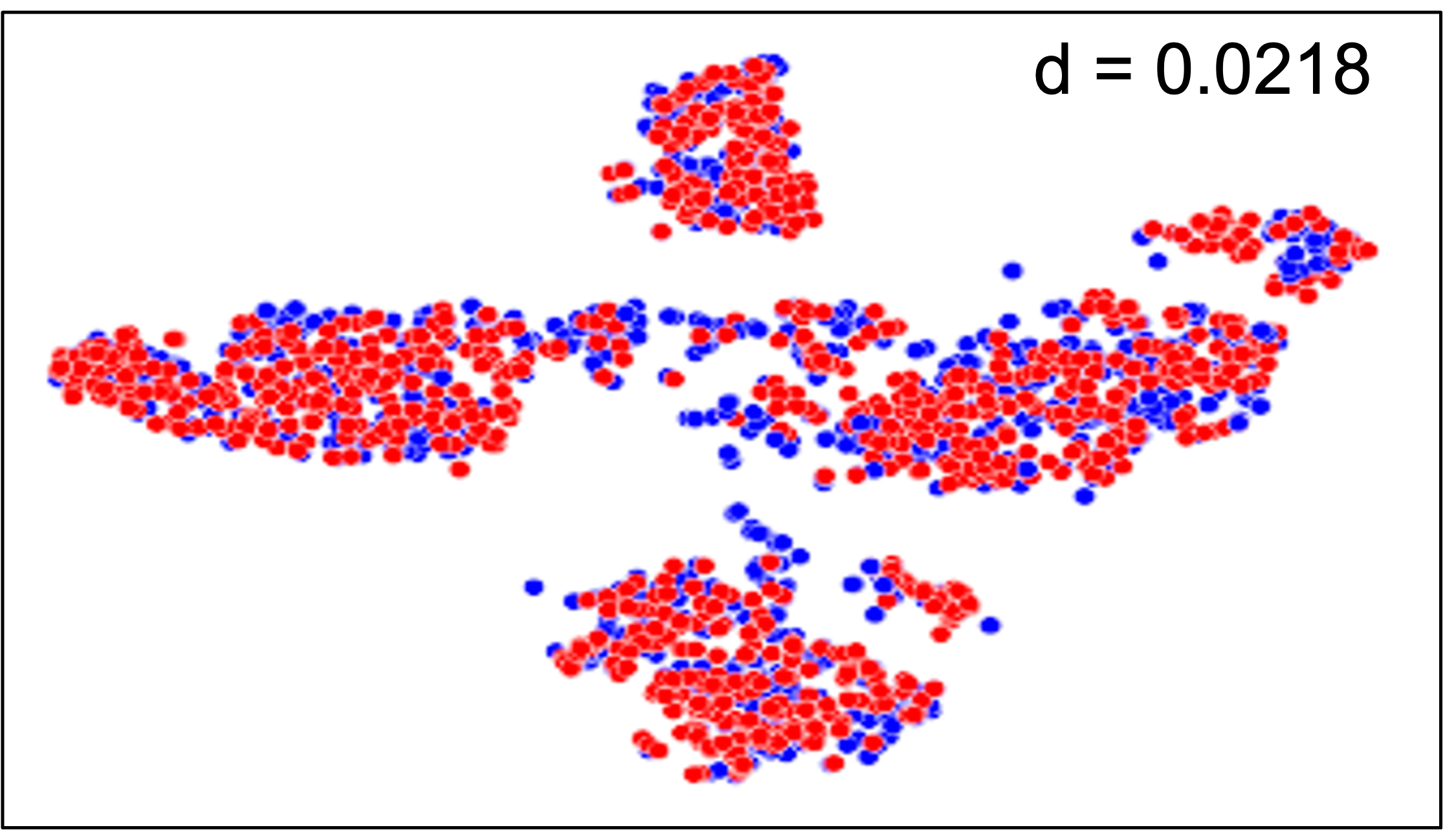}
    \caption{$ST_1$-transformed features}
  \end{subfigure}
  \begin{subfigure}{0.48\linewidth}
    \includegraphics[width=1.0\linewidth]{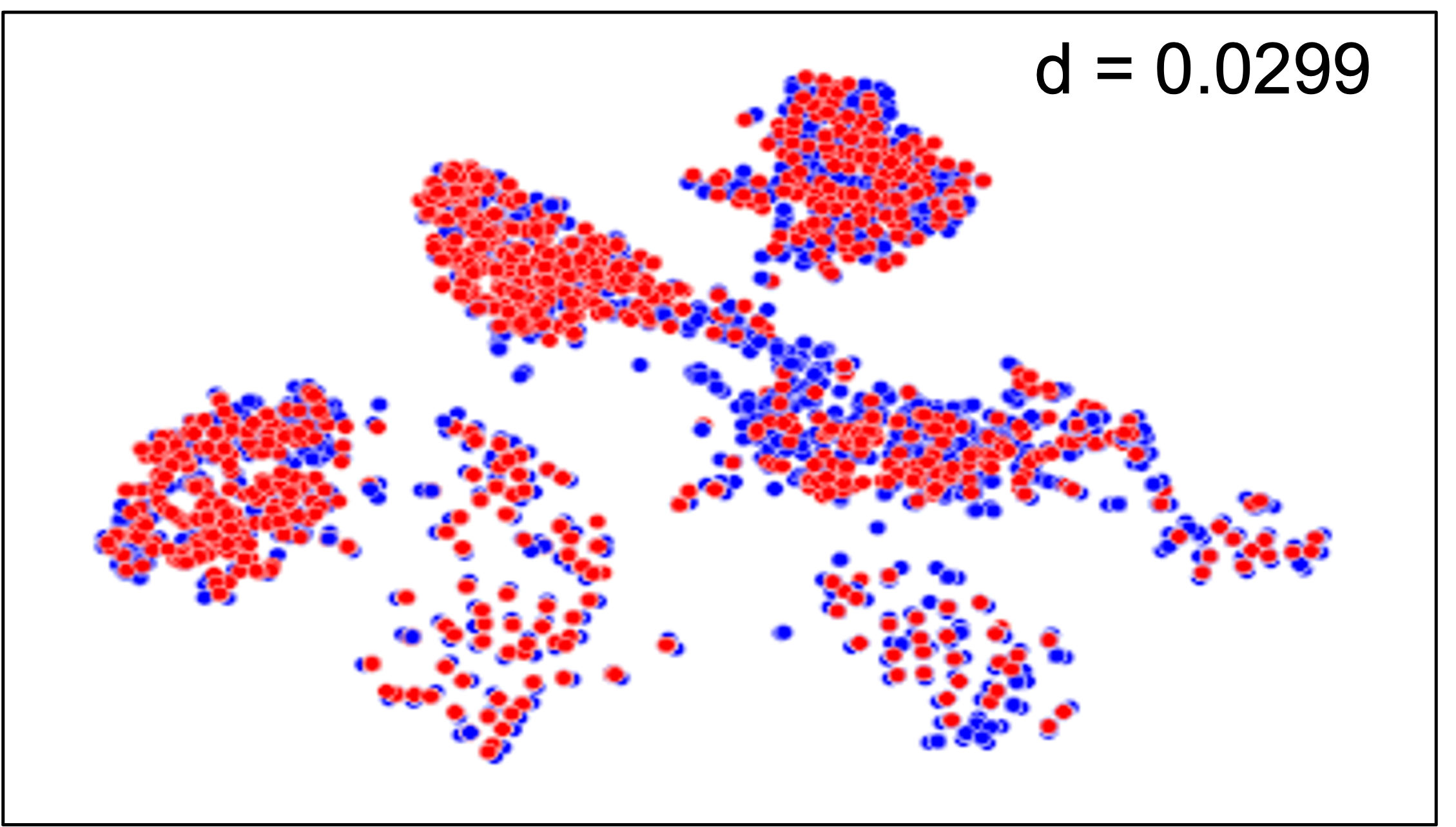}
    \caption{$ST_2$-transformed features}
  \end{subfigure}
  \hfill
  \begin{subfigure}{0.48\linewidth}
    \includegraphics[width=1.0\linewidth]{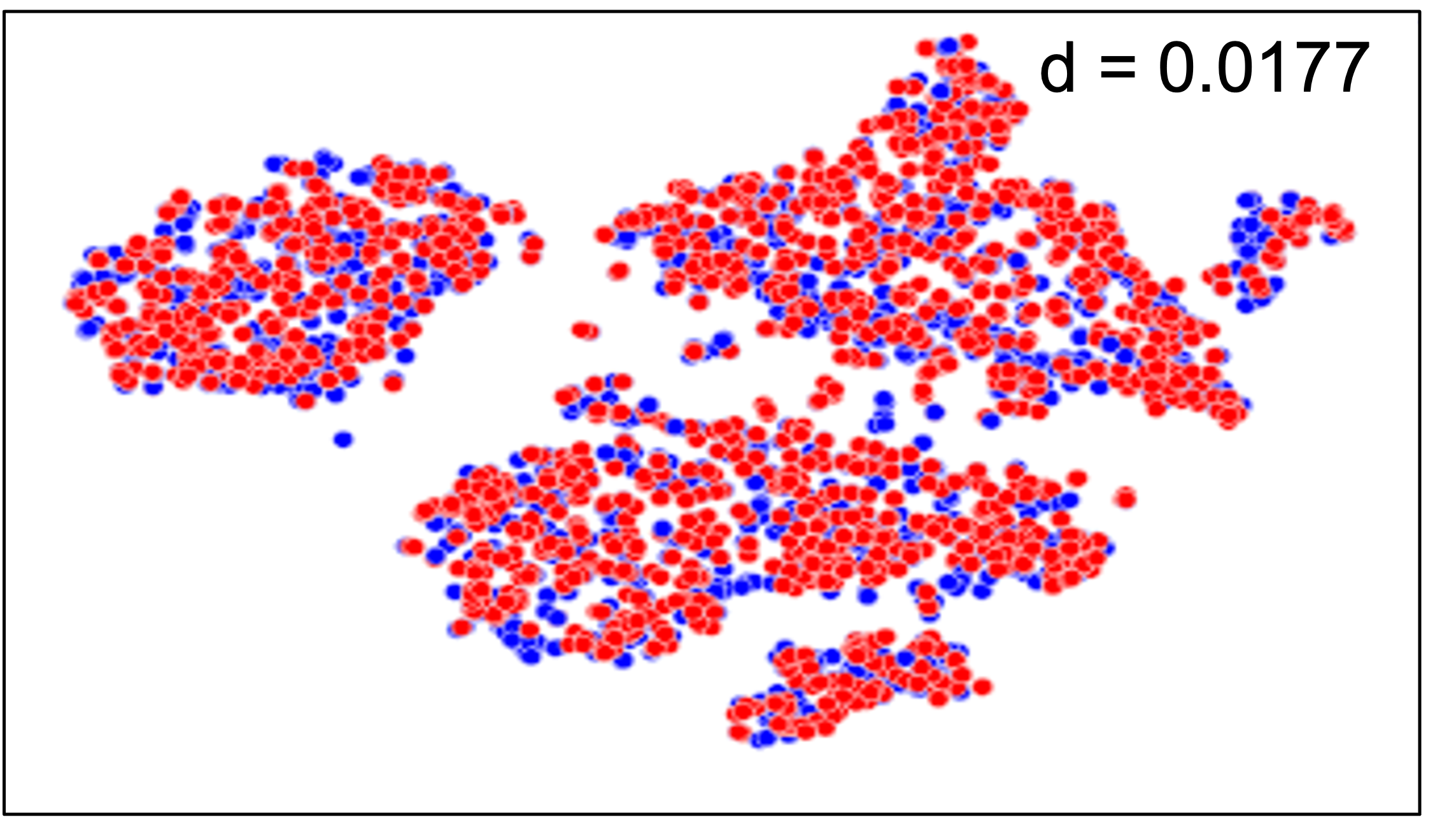}
    \caption{$(ST_1+ST_2)$-transformed features}
  \end{subfigure}
  \caption{ 
  Visualization of feature representations via t-SNE~\cite{maaten2008visualizing}: Red points represent source features and blue points represent target features.
  $d$ denotes the distance between source and target feature representations as measured by Maximum Mean Discrepancy~\cite{gretton2012mmd}. A single ST helps to reduce inter-domain distance significantly as shown in (b) and (c). Two complementary STs can further reduce the inter-domain distance clearly as shown in (d).
  }
\label{fig:tsne}
\end{figure}

\textbf{Complementarity Study:}
The proposed SUDA is complementary to most existing visual recognition methods consistently (while incorporated as a plug-in) as shown in Tables~\ref{table:det_city2fog}-\ref{table:synthia2city}. The synergistic effect is largely attributed to the proposed spectrum transformer and multi-view spectral learning which work in spectral domain whereas most existing methods work in spatial domain.

\textbf{ST Analysis:} 
We examine how the proposed ST learns to produce domain-invariant spectral information over domain adaptive object detection task Cityscapes $\rightarrow$ Foggy cityscapes. We first study the visual features of source and target samples that are produced by $ST_1$ and $ST_2$ in Fig.~\ref{fig:architecture}. As Fig.~\ref{fig:tsne} shows, the ST-generated source and target features are better aligned (with smaller inter-domain distances) as compared with that of the original images, showing that ST helps learn more domain-invariant features effectively. Due to space limit, we provide the visualization of ST-generated images in Section C.1 of the supplementary materials.

In addition, we analyze ST quantitatively by measuring the inter-domain distance~\cite{heusel2017fid} before and after the ST transformation of the source and target images. As Table~\ref{dis:FID_inter} shows, the cross-domain distance is reduced significantly by implementing either $ST_1$ (under ($X_s^1$, $X_t^1$)) or $ST_2$ (under ($X_s^2$, $X_t^2$)) alone as compared with the cross-domain distance of the original source and target samples. While $ST_1$ and $ST_2$ are both implemented, the cross-domain distance is further reduced as shown in the column ($X_s^{1,2}$, $X_t^{1,2}$).

\textbf{Analysis of Discrepancy Loss $\mathcal{L}_{dis}$:} We also study how $\mathcal{L}_{dis}$ guides $ST_1$ and $ST_2$ to learn more diverse features for both source and target samples. This study is based on intra-domain distance~\cite{heusel2017fid} over domain adaptive object detection task Cityscapes $\rightarrow$ Foggy cityscapes, $i.e.$, the larger the distance, the more diverse features learnt.

Specifically, we study the intra-domain distances of ST-generated source and target images while $\mathcal{L}_{dis}$ is present (as in ($X_s^1$, $X_s^2$) and ($X_t^1$, $X_t^2$)) and absent (as in (${X_s^1}'$, ${X_s^2}'$) and (${X_t^1}'$, ${X_t^2}'$)). 
As Table~\ref{dis:FID_intra} shows, the intra-domain distances are clearly larger when $\mathcal{L}_{dis}$ is present,  demonstrating that $\mathcal{L}_{dis}$ effectively guides the two STs to learn more diverse representations for both source and target samples.

\textbf{Number of STs:} 
The number of STs in SUDA does affect the domain adaptation performance. Our study shows that one ST can improve the domain adaptation significantly and two complementary STs can further introduce clear improvements.
However, the domain adaptation saturates with more STs which instead complicates the network structures and introduces extra parameters. Due to the space limit, we provide detailed experimental results and analysis in Section B.1 in supplementary materials.

\textbf{Comparisons with Existing Spectrum-based Techniques:} We compared SUDA with two existing spectrum-based UDA techniques~\cite{yang2020fda,huang2021rda}, where \cite{yang2020fda} swaps certain pre-defined FCs of source and target samples to mitigate inter-domain discrepancy whereas \cite{huang2021rda} employs adversarial attacking to mitigate the overfitting in UDA.  As a comparison, the proposed SUDA minimizes inter-domain discrepancy by identifying and enhancing domain-invariant FCs in a learnable manner. In addition, it introduces multi-view spectral learning for capturing more diverse target representations. SUDA thus addresses the UDA challenges from very different perspectives which is clearly complementary to the two spectrum-based works as shown in Table~\ref{table:gta2city}. We provide detailed comparison and analysis at Section B.2 of the supplementary materials.

\textbf{Parameter Analysis:}
We studied the sensitivity of the number of FCs $N$ and balance weights $\lambda_c$ and $\lambda_s$ defined in Eq.~\ref{all_loss} in Section B.4 in supplementary materials.

\renewcommand\arraystretch{1.0}
\begin{table}[t]
\centering
\begin{footnotesize}
\begin{tabular}{c|cccc}
\toprule
  &\multicolumn{1}{c}{${(X_s,X_t)}$}& \multicolumn{1}{c}{${(X_s^1,X_t^1)}$} &\multicolumn{1}{c}{${(X_s^2,X_t^2)}$} &\multicolumn{1}{c}{${(X_s^{1,2},X_t^{1,2})}$}  \\
\midrule
CDID& \multicolumn{1}{c}{58.57} & \multicolumn{1}{c}{18.78} & \multicolumn{1}{c}{20.56} & \multicolumn{1}{c}{16.23} \\
\bottomrule
\end{tabular}
\end{footnotesize}
\caption{
Quantitative analysis of ST on CDID (cross-domain image distance): The CDID (measured in FID~\cite{heusel2017fid}) is greatly reduced after transformation by either $ST_1$ in ($X_s^1$, $X_t^1$) or $ST_2$ in ($X_s^2$, $X_t^2$)
. The transformation by both $ST_1$ and $ST_2$ further reduces CDID clearly as in ($X_s^{1,2}$, $X_t^{1,2}$).
}
\label{dis:FID_inter}
\end{table}

\renewcommand\arraystretch{1.0}
\begin{table}[t]
\centering
\begin{footnotesize}
\begin{tabular}{c|cccc}
\toprule
&\multicolumn{1}{c}{${(X_s^1,X_s^2)}$} & \multicolumn{1}{c}{${(X_t^1,X_t^2)}$} &\multicolumn{1}{c}{${({X_s^1}',{X_s^2}')}$} &\multicolumn{1}{c}{${({X_t^1}',{X_t^2}')}$}  \\
\midrule
IDID&  \multicolumn{1}{c}{30.59} &  \multicolumn{1}{c}{29.82} &\multicolumn{1}{c}{7.13} &  \multicolumn{1}{c}{9.42}\\
\bottomrule
\end{tabular}
\end{footnotesize}
\caption{
Quantitative analysis of discrepancy loss $\mathcal{L}_{dis}$ on IDID (intra-domain image distance): With $\mathcal{L}_{dis}$, the IDID (measured in FID~\cite{heusel2017fid}) in ($X_s^1$, $X_s^2$) and ($X_t^1$, $X_t^2$) are clearly larger as compared with (${X_s^1}'$, ${X_s^2}'$) and (${X_t^1}'$, ${X_t^2}'$) without using $\mathcal{L}_{dis}$.
}
\label{dis:FID_intra}
\end{table}

\section{Conclusion}
This paper presents SUDA, a spectral UDA technique that addresses UDA challenges by learning domain-invariant spectral features. SUDA consists of two key designs. The first is a spectrum transformer that mitigates inter-domain discrepancy by highlighting domain-invariant spectra and suppressing domain-variant spectra in the input space. The second is multi-view spectral learning that leverage multi-view consistency for learning diverse representations for each target sample. SUDA has three unique features: 1) it is generic to various visual recognition tasks with consistently superior performance; 2) it is learnable and end-to-end trainable in various downstream tasks; 3) it complements with existing UDA methods with consistent performance boosts.
Moving forwards, we will continue to investigate frequency-space learning and its applications in various downstream computer vision tasks.

\textbf{Acknowledgement.} 
This study is supported under the RIE2020 Industry Alignment Fund – Industry Collaboration Projects (IAF-ICP) Funding Initiative, as well as cash and in-kind contribution from the industry partner(s). 
This research is also partially supported by Singtel Cognitive and Artificial Intelligence Lab for Enterprises (SCALE@NTU), which is a collaboration between Singapore Telecommunications Limited (Singtel) and Nanyang Technological University (NTU) that is supported by A*STAR under its Industry Alignment Fund (LOA Award number: I1701E0013).

{\small
\bibliographystyle{ieee_fullname}
\bibliography{egbib}
}

\end{document}